# Conditional independence in possibility theory

**Pascale FONCK**
Université de Liège
Institut de Mathématique
15, av. des Tilleuls
4000 Liège, BELGIUM
fonck@math.ulg.ac.be

## Abstract

Possibilistic conditional independence is investigated : we propose a definition of this notion similar to the one used in probability theory. The links between independence and no-interactivity are investigated, and properties of these relations are given. The influence of the conjunction used to define a conditional measure of possibility is also highlighted : we examine three types of conjunctions : Lukasiewicz - like T-norms, product-like T-norms and the minimum operator.

## 1 Introduction

Probabilistic expert systems include a representation of global knowledge under the form of a joint probability distribution, relative to a given set of variables.

In such expert systems, this joint distribution is the unique representation of the knowledge and every incoming piece of information is also modelled through a probability distribution. Belief updating in presence of new evidence consists in the combination of the knowledge and the incoming evidence, and the marginalization of this combination to each single variable.

When the number of variables is high, the joint probability distribution cannot easily be given to complete the knowledge-base.

To avoid this problem, one solution is to split the whole set of variables into small subsets, give probability distributions among those subsets, and then combine these distributions.

The notion of independence between sets of variables can be used to achieve this process.

Conditional probabilistic independence between sets of variables has been extensively studied (DAWID (1979)), and it has been shown that graphical structures, such as Markov networks (LAURITZEN & SPIEGELHALTER (1988)) or Bayesian networks (PEARL (1988)) provide a qualitative test for conditional independence. Moreover, the use of such structures allow to define "small" sets of variables on which probability distributions can be given and then combined using independence statements.

In the case of Markov networks, it suffices to specify probability distributions relative to the cliques of the network, and combine them into a single joint distribution (LAURITZEN & SPIEGELHALTER (1988)).

Using Bayesian networks, probability distributions on one node conditionally to its parents are combined to get the joint distribution (PEARL (1988)).

In both cases, a local algorithm of belief upadating is provided, whose aim is to propagate the effect of incoming evidence in the network (LAURITZEN & SPIEGELHALTER (1988), PEARL (1988)).

The algorithms exploit relations of independence displayed in the network and ignore irrelevant information.

The effectiveness of these algorithms is a great advantage for the use of probability theory in expert systems.

However, it has been shown that similar algorithms of belief updating are also available in different frameworks. SHAFER & SHENOY (1988) propose an algorithm of local computation of marginals, available in the framework of valuations, when the global knowledge factorizes on the edges of an acyclic hypergraph.

For such valuations, an algorithm of belief updating in directed acyclic graphs has been proposed by CANO, DELGADO & MORAL (1993).

In possibility theory, such algorithms of belief updating in hypergraphs (DUBOIS & PRADE (1990)) or in directed acyclic graphs (FONCK & STRASZECKA (1991)) can also be used.

Here we focus on possibility theory and provide a stu-dy of possibilistic conditional independence between sets of variables. The definition of possibilistic independence proposed here is different from the one introduce by ZADEH, called no-interactivity. We use it as an irrelevance relation between pieces of information.

Conditional independence relations are introduced in section 2 and possibility theory is recalled in section 3.

Possibilistic conditional independence relations are then defined in section 4 and compared with the relations of no-interactivity. The properties of these relations are drawn. Section 5 brings the conclusion.

## 2 Conditional independence relations

In this paper, we will focus on independence between sets of variables, rather than on independence between events.

Let us denote by $\mathbb{X}$ a finite set of variables, taking values in finite sets. The frame associated to variable $X$ is denoted by $\Omega_X$, and throughout this paper, the notation $\Omega_A$ stands for $(\underset{X \in A}{X} \Omega_A)$, with $A \subseteq \mathbb{X}$.

Let $T(\mathbb{X})$ denote the set of triplets of pairwise disjoint subsets of $\mathbb{X}$.

**Definition 1** A *conditional independence relation* on $\mathbb{X}$ is a subset $I$ of $T(\mathbb{X})$. When $(A, B, C) \in I$, $A$ is said independent from $B$, conditionally to $C$, for the relation $I$.

Some conditional independence relations play a central role in inference networks theory.

**Definition 2** A *semigraphoid* on $\mathbb{X}$ is a conditional independence relation $I$ on $\mathbb{X}$ fulfilling the following properties:

- symmetry : $(A, B, C) \in I \Rightarrow (B, A, C) \in I$
- decomposition :
$$(A, B \cup C, D) \in I \Rightarrow (A, B, D) \in I$$
- weak union :
$$(A, B \cup C, D) \in I \Rightarrow (A, B, C \cup D) \in I$$
- contraction :
$$\left.\begin{array}{l}(A, B, D) \in I \\ (A, C, B \cup D) \in I\end{array}\right\} \Rightarrow (A, B \cup C, D) \in I$$

A *graphoid* on $\mathbb{X}$ is a semigraphoid $I$ on $\mathbb{X}$ fulfilling also the property :

- intersection :
$$\left.\begin{array}{l}(A, B, C \cup D) \in I \\ (A, C, B \cup D) \in I\end{array}\right\} \Rightarrow (A, B \cup C, D) \in I$$

Graphoids are conditional independence relations which can be uniquely represented through an inference network, i.e. a directed acyclic graph whose nodes are the elements of $\mathbb{X}$ and links express relations of dependence.

Semigraphoids can also be represented by such networks, but this representation is not unique. Probabilistic independence relations induced by strictly positive probability distributions are graphoids : this result was used by PEARL (1988) to propose a method to build a probabilistic knowledge-base from local informations and to achieve an inference process, based on Bayesian networks.

Other types of graphoids are also encountered : see for example HUNTER (1991).

In this paper, we are mainly interested by possibilistic conditional independence relations. Let us first settle some basic notions about possibility distributions.

## 3 Possibility distributions

Possibility distributions can be used to model uncertain pieces of information about the elements in $\mathbb{X}$.

**Definition 3** A normalised possibility distribution on $\Omega_{\mathbb{X}}$ is an application $\pi$ from $\Omega_{\mathbb{X}}$ into $[0,1]$, such that
$$\exists x \in \Omega_{\mathbb{X}} : \pi(x) = 1.$$

Let us note that a normalised possibility distribution over a finite set $\Omega_{\mathbb{X}}$ induces a possibility measure on $(\Omega_{\mathbb{X}}, P(\Omega_{\mathbb{X}}))$, defined by
$$\Pi(A) = \sup_{x \in A} \pi(x), \forall A \in P(\Omega_{\mathbb{X}}).$$

In an equivalent way, a possibility distribution $\pi$ over $\Omega_{\mathbb{X}}$ can be seen as the membership function of a fuzzy set.

For any $x \in \Omega_{\mathbb{X}}$, the number $\pi(x)$ represents the degree of possibility of the event "the multidimentional variable $(X)_{X \in \mathbb{X}}$ takes the value $x$", or the degree of membership of $x$ to the fuzzy subset of the possible values of the multidimensional variable $(X)_{X \in \mathbb{X}}$.

Two basic operations on possibility distributions are defined.

The operation of marginalization consists in the extraction from a body of knowledge about the elements in $\mathbb{X}$ of some information about a subset of $\mathbb{X}$.

**Definition 4** The *marginalization* of the possibility distribution $\pi$ on $\Omega_{\mathbb{X}}$ to $A \subseteq X$ is the possibility distribution $\pi_A$ defined on $\Omega_A$ by
$$\pi_A(x_A) = \max_{y \in \Omega_{\mathbb{X} \setminus A}} \pi(x_A, y), \forall x_A \in \Omega_A.$$

This distribution $\pi_A$ is the least restrictive distribution defined on $\Omega_A$, compatible with $\pi$. The operation of marginalization obeys of course the consonance property.

**Property 1** If $B \subseteq A \subseteq \mathbb{X}$ and $\pi$ is a possibility distribution over $\Omega_{\mathbb{X}}$, then $(\pi_A)_B = \pi_B$.

Conditional possibility distributions can also be computed from a joint distribution.

Such distributions are used to express the relationships between different subsets of $\mathbb{X}$. The definition of these distributions is based on the choice of an operation of conjunction.

**Definition 5** An *operation of conjunction* is a function $c$ from $[0,1] \times [0,1]$ into $[0,1]$, continuous, non decreasing in both arguments, and such that

$$c(0,a) = c(a,0) = 0, \forall a \in [0,1]$$
$$c(1,a) = c(a,1) = a, \forall a \in [0,1].$$

**Definition 6** If $A$ and $B$ are disjoint subsets of $\mathbb{X}$ and $\pi$ is a normalised possibility distribution over $\Omega_{\mathbb{X}}$, then the *possibility distribution over $A$ conditionally to $B$* induced by $\pi$ is

$$\pi_{A|B} : \Omega_{A \cup B} \to [0,1]$$
$$: (x_A, x_B) \to \Im_c(\pi_B(x_B), (\pi_{A \cup B}(x_A, x_B)),$$

with

$$\Im_c(a, b) = \sup\{s \in [0,1] : c(s, a) \leq b\}.$$

Examples of conditional possibility distributions are given below.

**Example 1** If $c$ denotes a Lukasiewicz - like $T$ - norm, i.e. if $c(a,b) = \varphi^{-1}(Tm(\varphi(a), \varphi(b)))$, with

$$Tm(a, b) = max(0, a + b - 1), \forall a, b \in [0,1]$$

and $\varphi$ denotes a continuous strictly increasing function from $[0,1]$ into $[0,1]$, such that $\varphi(0) = 0$ and $\varphi(1) = 1$; conditional possibility distributions are defined by :

$$\pi_{A|B}^{\varphi(Tm)}(x_A, x_B)$$
$$= \varphi^{-1}(\varphi(\pi_{A \cup B}(x_A, x_B)) - \varphi(\pi_B(x_B)) + 1).$$

**Example 2** If $c$ denotes a product-like $T$-norm, i.e. if $c(a,b) = \varphi^{-1}(\varphi(a)\varphi(b))$, conditional possibility distributions are defined by

$$\pi_{A|B}^{\varphi(P)}(x_A, x_B)$$
$$= \begin{cases} \varphi^{-1}(\frac{\varphi(\pi_{A \cup B}(x_A, x_B))}{\varphi(\pi_B(x_B))}) & \text{if } \pi_B(x_B) > 0 \\ 1 & \text{else} \end{cases}$$

**Example 3** If $c = \min$, conditional possibility distributions are defined by

$$\pi_{A|B}^{min}(x_A, x_B)$$
$$= \begin{cases} \pi_{A \cup B}(x_A, x_B) & \text{if } \pi_{A \cup B}(x_A, x_B) < \pi_B(x_B) \\ 1 & \text{else} \end{cases}$$

These three types of conjunctions will be examined in the following. Whatever the conjunction $c$ used, the identity

$$c(\pi_{A|B}, \pi_B) = \pi_{A \cup B}$$

is of course true, for any distribution $\pi$. However, the distribution $\pi_{A \cup B}$ cannot be recovered from $\pi_A$ and $\pi_B$, unless independence statements are expressed.

Independence relations are introduced in the next section.

## 4 Possibilistic conditional independence relations

Let $\pi$ be a normalised possibility distribution over $\Omega_{\mathbb{X}}$.

By analogy with probability theory, the conditional independence relation induced by $\pi$ on $\mathbb{X}$ is based on conditional distributions induced by $\pi$.

**Definition 7** The *conditional independence relation* $I_\pi^c$ induced by $\pi$ is defined as :

$$I_\pi^c = \{(A, B, C) \in T(\mathbb{X}) : \pi_{A|B \cup C}^c = \pi_{A|C}^c$$
$$\text{and } \pi_{B|A \cup C}^c = \pi_{B|C}^c\}.$$

Let us note that in possibility theory, the equivalence

$$\pi_{A|B \cup C}^c = \pi_{A|C}^c \Leftrightarrow \pi_{B|A \cup C}^c = \pi_{B|C}^c$$

is not fulfilled, as can be seen with the following example.

**Example 4** Let $\mathbb{X} = \{X_1, X_2, X_3\}$, $\Omega_{X_1} = \Omega_{X_2} = \Omega_{X_3} = \{0,1\}$, and

$$\pi(x_1, x_2, x_3) = \begin{cases} 0.6 & \text{if } x_1 = 0 \\ 0.7 & \text{if } x_1 = 1, x_2 = 0, x_3 = 0 \\ 0.8 & \text{if } x_1 = 1, x_2 = 0, x_3 = 1 \\ 0.9 & \text{if } x_1, x_2 = 1, x_3 = 0 \\ 1 & \text{if } x_1 = x_2 = x_3 = 1 \end{cases}$$

Then

$$\pi_{\{X_1\}|\{X_2, X_3\}}^{min} = \pi_{\{X_1\}|\{X_3\}}^{min}$$

and

$$\pi_{\{X_2\}|\{X_1, X_3\}}^{min} \neq \pi_{\{X_2\}|\{X_3\}}^{min}$$

This conditional independence relation can be interpreted in the following way : the sets $A$ and $B$ are independent, conditionally to $C$ if and only if once the values of the elements of $C$ are known, further information about the elements of $B$ is irrelevant for $A$, and further information about the elements of $A$ is irrelevant for $B$.

This notion of independence is stronger than the one commonly used in possibility theory, which was introduced by ZADEH.

This notion, called "no-interactivity" is defined as: $A$ and $B$ are not interacting on each other, conditionally to $C$, for the distribution $\pi$, if and only if $\pi_{A \cup B|C}$ factorizes into $\min(\pi_{A|C}, \pi_{B|C})$, where the conjunction used is the minimum operator. This can be extended to the case of any conjunction.

**Definition 8** The *relation of no-interactivity* $NI_\pi^c$ induced by $\pi$ on $\mathbb{X}$ is defined as:

$$NI_\pi^c = \{(A, B, C) \in T(\mathbb{X}) : \pi_{A \cup B|C}^c = c(\pi_{A|C}^c, \pi_{B|C}^c)\}.$$

Having two different independence relations, we now examine them in three particular cases : Lukasiewicz - like $T$-norms, product - like $T$-norms and minimum operator.

### 4.1 Lukasiewicz - like $T$ - norms

The following result shows the equivalence between conditional independence and no-interactivity, when a Lukasiewicz - like $T$-norm $\varphi(Tm)$ is used as a conjunction.

**Property 2** If $\pi$ is a normalised possibility distribution on $\Omega_{\mathbb{X}}$, then

$$I_\pi^{\varphi(Tm)} = NI_\pi^{\varphi(Tm)}$$

and $(A, B, C) \in I_\pi^{\varphi(Tm)}$ if and only if

$$\varphi(\pi_{A \cup B \cup C}) + \varphi(\pi_C) = \varphi(\pi_{A \cup C}) + \varphi(\pi_{B \cup C}).$$

Furthermore, the following result provides a sufficient condition for a triplet $(A, B, C)$ to be an element of $I_\pi^{\varphi(Tm)}$

**Property 3** If there exist possibility distributions $f_{A \cup C}$ and $f_{B \cup C}$, respectively defined on $\Omega_{A \cup C}$ and $\Omega_{B \cup C}$, such that

$$(f_{A \cup C})_C = (f_{B \cup C})_C,$$

$$\pi_{A \cup B \cup C} = \varphi(Tm)(f_{A \cup C}, f_{B \cup C}),$$

and $\varphi(f_{A \cup C}) + \varphi(f_{B \cup C}) \geq 1$, then $(A, B, C) \in I_\pi^{\varphi(Tm)}$.

**Property 4** The relation $I_\pi^{\varphi(Tm)}$ is a graphoid.

This last property shows that the relation $I_\pi^{\varphi(Tm)}$ can be uniquely represented by an inference network.

**Remark** : Using a Lukasiewicz - like $T$ - norm as an operation of conjunction, we get a conditional measure of possibility $\Pi(./A)$ such that $\Pi(\overline{A}/A)$ is in general different from 0. If this property, expressing the incompatibility between a crisp event and its contrary, seems necessary for the user, the Lukasiewicz - like $T$ - norms should be neglected to define a conditional possibility measure, to the benefit of another conjunction, such as the minimum operator, or a product - like $T$ - norm.

### 4.2 Product-like $T$-norms

When a product - like $T$-norm $\varphi(P)$ is used as a conjunction to define conditional possibility measures, the relations of independence and no-interactivity are no longer equivalent, as stated in the following result.

**Property 5** Let $\pi$ be a normalised possibility distribution over $\Omega_{\mathbb{X}}$ and $(A, B, C) \in T(\mathbb{X})$.

Then

$(A, B, C) \in NI_\pi^{\varphi(P)}$ if and only if

$$\varphi(\pi_{A \cup B \cup C})\varphi(\pi_C) = \varphi(\pi_{A \cup C})\varphi(\pi_{B \cup C}).$$

Moreover, $(A, B, C) \in I_\pi^{\varphi(P)}$ if and only if

$$(A, B, C) \in NI_\pi^{\varphi(P)}$$

and

$\forall x_C \in \Omega_C$ such that $\pi_C(x_C) > 0$, if there exists $x_B \in \Omega_B : \pi_{B \cup C}(x_B, x_C) = 0$, then

$$\pi_{A \cup C}(x_A, x_C) = \pi_C(x_C), \forall x_A \in \Omega_A$$

and

$\forall x_C \in \Omega_C$ such that $\pi_C(x_C) > 0$, if there exists $x_A \in \Omega_A : \pi_{A \cup C}(x_A, x_C) = 0$, then

$$\pi_{B \cup C}(x_B, x_C) = \pi_C(x_C), \forall x_B \in \Omega_B.$$

The properties of $I_\pi^{\varphi(P)}$ and $NI_\pi^{\varphi(P)}$ are listed below.

**Property 6** The relation $I_\pi^{\varphi(P)}$ is a graphoid. The relation $NI_\pi^{\varphi(P)}$ is a semigraphoid.

Let us note that $NI_\pi^{\varphi(P)}$ is not a graphoid : this can be seen on the following example.

**Example 5** Let $\mathbb{X} = \{X_1, X_2, X_3\}$, $\Omega_{X_1} = \{0, 2\}, \Omega_{X_2} = \{-1, 1\} = \Omega_{X_3}$, and

$$\pi(x_1, x_2, x_3) = \begin{cases} 1 \text{ if } x_1 = 0, x_2 = 1, x_3 = -1 \\ 1 \text{ if } x_1 = 2, x_2 = -1, x_3 = 1 \\ 0 \text{ else} \end{cases}$$

Then

$$(\{X_1\}, \{X_2\}, \{X_3\}) \in NI_\pi^{\varphi(P)}$$
$$(\{X_1\}, \{X_3\}, \{X_2\}) \in NI_\pi^{\varphi(P)},$$

and

$$(\{X_1\}, \{X_2, X_3\}, \emptyset) \in NI_\pi^{\varphi(P)}.$$

We thus get a notion of no-interactivity weaker than the one of conditional independence. Let us note that, if $\pi$ is a strictly positive possibility distribution, then

$$I_\pi^{\varphi(P)} = NI_\pi^{\varphi(P)},$$

and the results about $I_\pi^{\varphi(P)}$ are quite similar to the results about probabilistic independence relations.

### 4.3 Minimum operator

When the minimum operator is used as a conjunction, the independence property can be expressed as a property of the joint distribution.

**Property 7** Let $\pi$ be a normalised possibility distribution over $\Omega_{\mathbb{X}}$ and $(A, B, C) \in T(\mathbb{X})$. Then

$$(A, B, C) \in I_\pi^{\min} \text{ if and only if}$$
$$\pi_{A \cup B \cup C} = \min(\pi_{A \cup C}, \pi_{B \cup C})$$

and

$$\pi_C = \max(\pi_{A \cup C}, \pi_{B \cup C}).$$

Moreover, $(A, B, C) \in NI_\pi^{\min}$ if and only if

$$\pi_{A \cup B \cup C} = \min(\pi_{A \cup C}, \pi_{B \cup C}).$$

As in the previous case, no-interactivity is weaker than conditional independence.

**Property 8** The relation $I_\pi^{\min}$ is a graphoid. The relation $NI_\pi^{\min}$ is a semigraphoid.

Example 5 shows that $NI_\pi^{\min}$ is not a graphoid : it lacks the intersection property.

## 5 Conclusion

In this paper, we have considered two types of possibilistic independence relations : conditional independence relations and no-interactivity relations.

These two notions were shown to be in general not equivalent, and their properties were listed.

Conditional independence relations are graphoids : they can thus be used for the building of a Markov network or an inference network to model an uncertain knowledge-base.

No-interactivity relations in general lack the intersection property : these relations cannot be used for the building of a Markov network and their representation through an inference network is not unique.

Moreover, conditional independence can be interpreted in terms of relevance of information. That is why we have chosen to use possibilistic conditional independence for the modelling of an uncertain knowledge-base through an inference network, and local algorithms of belief revision are then available (see FONCK (1993)).